\documentclass{bmvc2k}

\usepackage{amsthm}
\usepackage{booktabs}
\usepackage{algorithm}
\usepackage{algorithmic}
\usepackage{xspace}
\usepackage{amsfonts,amssymb}
\usepackage{multirow}
\usepackage{subfig}
\usepackage{graphicx}
\usepackage{wrapfig}

\title{Robust Image Matching \\By Dynamic Feature Selection}

\addauthor{Hao Huang}{hh1811@nyu.edu}{123}
\addauthor{Jianchun Chen}{jc7009@nyu.edu}{12}
\addauthor{Xiang Li}{xl1845@nyu.edu}{123}
\addauthor{Linging Wang}{lingjing.wang@nyu.edu}{123}
\addauthor{Yi Fang$^\ast$}{yfang@nyu.edu}{123}

\addinstitution{
 NYU Multimedia and Visual \\Computing Lab, USA
}

\addinstitution{
 New York University, USA
}
\addinstitution{
 New York University Abu Dhabi, UAE
}

\runninghead{Huang, Chen, Li, Wang, Fang}{Image Matchng \& Feature Selection}

\def\eg{\emph{e.g}\bmvaOneDot}

\def\etal{\emph{et al}\bmvaOneDot}

\newcommand{\head}[1]{\noindent\textbf{#1}}

\DeclareMathOperator*{\argmax}{argmax}

\makeatletter
\DeclareRobustCommand\onedot{\futurelet\@let@token\@onedot}
\def\@onedot{\ifx\@let@token.\else.\null\fi\xspace}

\def\ie{\emph{i.e}\onedot}

\makeatother

\begin{document}

\maketitle

\begin{abstract}
Estimating dense correspondences between images is a long-standing image understanding task. Recent works introduce convolutional neural networks (CNNs) to extract high-level feature maps and find correspondences through feature matching. However, high-level feature maps are in low spatial resolution and therefore insufficient to provide accurate and fine-grained features to distinguish intra-class variations for correspondence matching. To address this problem, we generate robust features by dynamically selecting features at different scales. To resolve two critical issues in feature selection, \ie, \textit{how many} and \textit{which} scales of features to be selected, we frame the feature selection process as a sequential Markov decision-making process (MDP) and introduce an optimal selection strategy using reinforcement learning (RL). We define an RL environment for image matching in which each individual action either requires new features or terminates the selection episode by referring a matching score. Deep neural networks are incorporated into our method and trained for decision making. Experimental results show that our method achieves comparable/superior performance with state-of-the-art methods on three benchmarks, demonstrating the effectiveness of our feature selection strategy.
\end{abstract}

\section{Introduction}
Image matching is one of the fundamental image understanding problems and serves as a building block for various applications, \ie, object recognition~\cite{liu2008sift}, motion tracking~\cite{newcombe2011dtam}, and 3D construction~\cite{agarwal2011building}. Image matching techniques can be further extended or adapted to other application fields including remote sensing~\cite{li2018building}, protein analysis~\cite{liu2009using} and medical imaging~\cite{audette2000algorithmic}. Existing methods tackle image matching problem by fitting a geometric transformation between the correspondences of image pairs. Previous approaches usually consist of two steps: 1) a group of hand-crafted image descriptors, \eg, HOG, SIFT, or SURF, are pre-computed to obtain pixel-level descriptions; 2) feature matching algorithms, \eg, RANSAC or Hough transformation, are then applied in an iterative manner to determine a proper geometric transformation to match pixels using their associated features. However, hand-crafted image descriptors are vulnerable to intra-class variations, \eg, different texture, lighting, and material. Recently, learning-based approaches~\cite{han2017scnet,rocco2017convolutional} alleviate this problem by extracting image features through CNNs and treat image feature maps as dense image descriptors. In general, learning-based methods can be divided into two categories. Some works~\cite{rocco2017convolutional,hongsuck2018attentive} focus on predicting parameters of a global rigid or non-rigid transformation, \eg, an affine or a thin-plate spline (TPS) transformation. Others~\cite{han2017scnet,rocco2018end} formulate this task as a local region matching process and directly pair local regions in source images to the matched regions in target images without transformation regression.

These methods, however, generate dense correspondences only based on high-level features, \ie, the outputs of the last and/or penultimate convolutional layers~\cite{rocco2017convolutional,rocco2018end}. As features produced by different levels of CNN layers contain information varying from low-level texture patterns to high-level semantic concepts, these methods fail to fully exploit multi-level image abstractions to obtain reliable matches that are robust to intra-class variations and background noise. 
\begin{figure}
\centering
  \includegraphics[width=.8\linewidth]{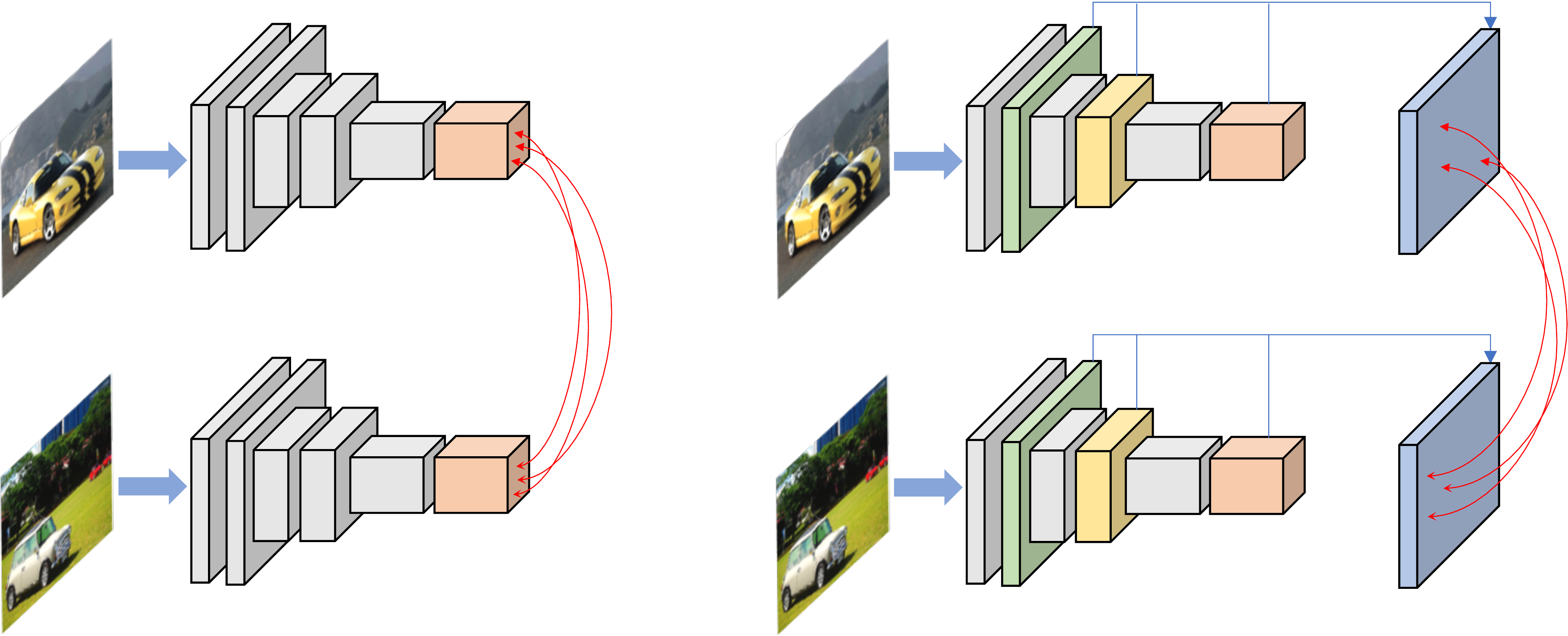}
  \vspace{5pt}
  \caption{Comparison between image matching methods that only use high-level features (left) and ours that select and integrate features at multiple scales (right).}
  \label{fig:teaser}
\end{figure}
The comparison between image matching methods that only use high-level features and multi-scale features is shown in Figure~\ref{fig:teaser}. 
However, there are two pivotal issues in utilizing different levels of features: \textit{how many} and \textit{which} levels of features are apt for matching. To address the issues, the previous method~\cite{min2019hyperpixel} applies beam search to select features from low levels to high levels sequentially without jumping. However, beam search is a heuristic searching strategy, and the search space is a subspace of the full solution space. If multiple optimal solutions exist, beam search may fail to find the one satisfying other requirements, \ie, a minimal number of features used for matching. In this paper, we frame the feature selection problem as a sequential Markov decision-making process (MDP) and tackle it using reinforcement learning. Specifically, based on the selected features, each individual action either requires new features or terminates the selection episode by referring a matching score. The learning process is driven by reward functions. Without manually imposed prior knowledge about image pairs, the proposed method can select the optimal collection of features that are suitable for image matching. Compared with beam search, there is no strict selection order in our proposed method, \ie, from low to high levels, leading to a larger search space and a higher possibility to find the optimal solution. We test the proposed method on three public datasets to demonstrate the effectiveness of our proposed feature selection strategy for robust image matching. Our paper makes three main contributions:
\begin{enumerate}
    \item We are the first to cast the feature selection process as an MDP and adopt reinforcement learning to select multiple levels of features for robust image matching.
    \item We devise a simple but effective deep neural networks to fuse selected features at multiple levels and make a decision at each step, \ie, either to select a new feature or to stop selection for evaluation.
    \item We achieve superior/comparable results on three public benchmarks for image semantic correspondence estimation, demonstrating the effectiveness of our method.
\end{enumerate}

\section{Related Work}
\head{Local region matching.} Methods in this category match two sets of local regions based on feature similarity. Traditional methods apply hand-crafted features with spatial regularization~\cite{hur2015generalized} or random sampling~\cite{barnes2009patchmatch}. Due to the sparsity and vulnerability of the hand-crafted features, they are incapable of handling images containing complex scenes. Bristow \etal~\cite{bristow2015dense} use LDA-whitened SIFT descriptors, making correspondences less vulnerable to background clutters. Cho \etal~\cite{cho2015unsupervised} introduces an effective voting-based algorithm based on region proposals and HOG features for semantic matching and object discovery. Ham \etal~\cite{ham2017proposal} extends~\cite{cho2015unsupervised} with a local-offset matching algorithm. Recently, features extracted by CNNs have replaced hand-crafted features. 
Features produced by CNNs can represent high-level semantics and are robust to appearance and shape variations~\cite{deng2009imagenet,dai2018siamese,xie2016learned}. Choy \etal~\cite{choy2016universal} proposes a similarity metric based on CNN features using a contrastive loss. Rocco \etal~\cite{rocco2018neighbourhood,lee2019sfnet} trains image features and correlation features for correspondence matching in an unsupervised/self-supervised setting. Min \etal~\cite{min2019hyperpixel} generates a pixel flow to match local regions via a Hough Voting procedure. Similarly, we also adopt CNN features for image matching. 

\head{Global image alignment.} Other methods~\cite{rocco2017convolutional,hongsuck2018attentive} formulate correspondence estimation as a geometric alignment problem. Parametric models are adopted to regress parameters of a global transformation. Specifically, image correlation tensors built on image features are fed into a regression layer/network to predict transformation parameters. Some works~\cite{rocco2017convolutional,hongsuck2018attentive} utilize synthetic image pairs, while the work~\cite{rocco2018end} explores to match images in a weakly supervised way. In addition, Kim \etal~\cite{kim2018recurrent} introduces a recurrent spatial transformer to apply local affine transformations to images iteratively. Chen \etal~\cite{chen2019arbicon} propose a global transformation independent of affine or TPS assumptions. However, these methods only apply high-level features, \ie, the last and/or penultimate features, generated by CNNs to construct correlation tensors, ignoring the importance of exploiting multi-scale levels of features.

\head{Reinforcement Learning in Vision.} Our method is largely inspired by works on leveraging reinforcement learning to solve vision problems.~\cite{mnih2014recurrent} is a classical work that uses RL for the spatial attention policy in image recognition. RL is also adopted for object detection~\cite{pirinen2018deep}, video object segmentation~\cite{han2018reinforcement}, video recognition~\cite{wu2019multi}, object tracking in~\cite{ren2018collaborative}, scene completion~\cite{han2019deep} and point cloud parsing~\cite{liu20173dcnn}. 
Our work is the first attempt to explore using RL to select a portion of hierarchical features produced by CNNs in the image matching scenario.

\section{Approach}
In this section, we detail our reinforcement learning method for image matching: instantiations of three basic concepts (\ie, state, action, reward) in reinforcement learning (Sec.~\ref{subsec:overall}), network architecture fusing selected features at multiple abstraction levels (Sec.~\ref{subsec:qnet}) and reinforcement training process (Sec.~\ref{subsec:rl}). An overview of our method is shown in Figure~\ref{fig:arch}.

\subsection{Problem Formulation}
Given an input image $I$, a backbone CNN with $N$ convolutional layers produces a sequence of $N$ features $S=(f_1,f_1,\cdots,f_{N})$ as intermediate representations, ranging from low level visual patterns to high level object semantics. We need to select a subset of rich and compact features for image matching. Once a collection of features are selected, we adopt the probabilistic Hough matching~\cite{min2019hyperpixel} to match images based on the selected features. A brief description of probabilistic Hough matching can be found in Appendix. We formulate the feature selection as a problem of maximizing the matching score function $f_m:\mathbb{S}\rightarrow R$ over the power set $\mathbb{S}$ of the feature set $S$:
\begin{align}
    s^\ast = \argmax_{s\in \mathbb{S} \land s \neq \emptyset}f_m(s)\enspace.
\end{align}
We treat the selection process as a Markov decision process, which is applicable of modeling the discrete sequential decision making process. To reframe MDP in image matching scenario, a state set $\mathcal{S}$, an action set $\mathcal{A}$, and a reward function $\mathcal{R}$ are defined as follows.

\subsection{Overall Architecture}
\label{subsec:overall}
\begin{figure}
\centering
\begin{minipage}{.48\textwidth}
  \centering
  \includegraphics[width=1\linewidth]{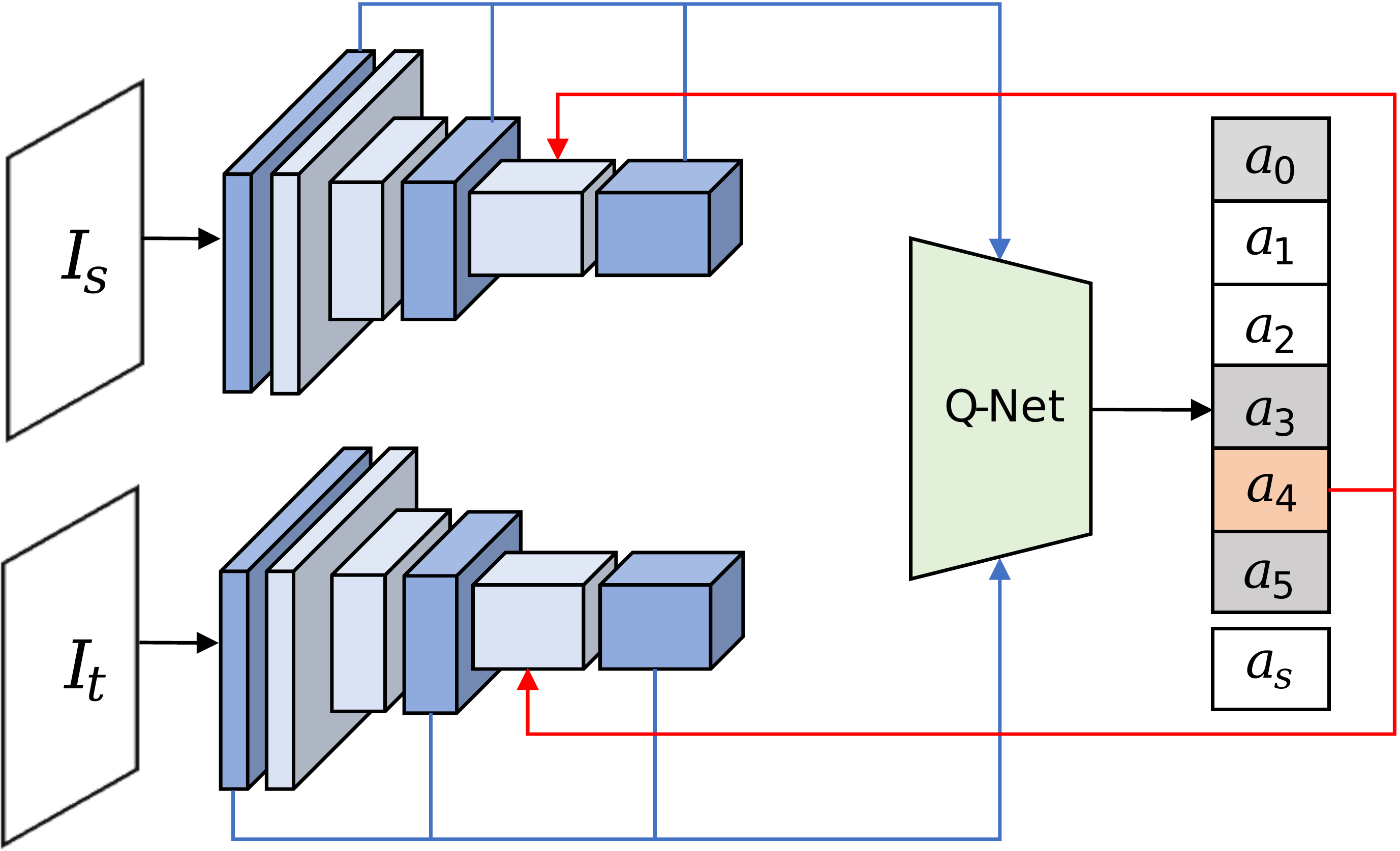}
  \vspace{5pt}
  \captionof{figure}{An overview of the proposed approach. Actions in grey color are invalid actions at the current step, as the corresponding features have been selected previously.}
  \label{fig:arch}
\end{minipage}\enspace%
\begin{minipage}{.48\textwidth}
  \centering
  \includegraphics[width=.94\linewidth]{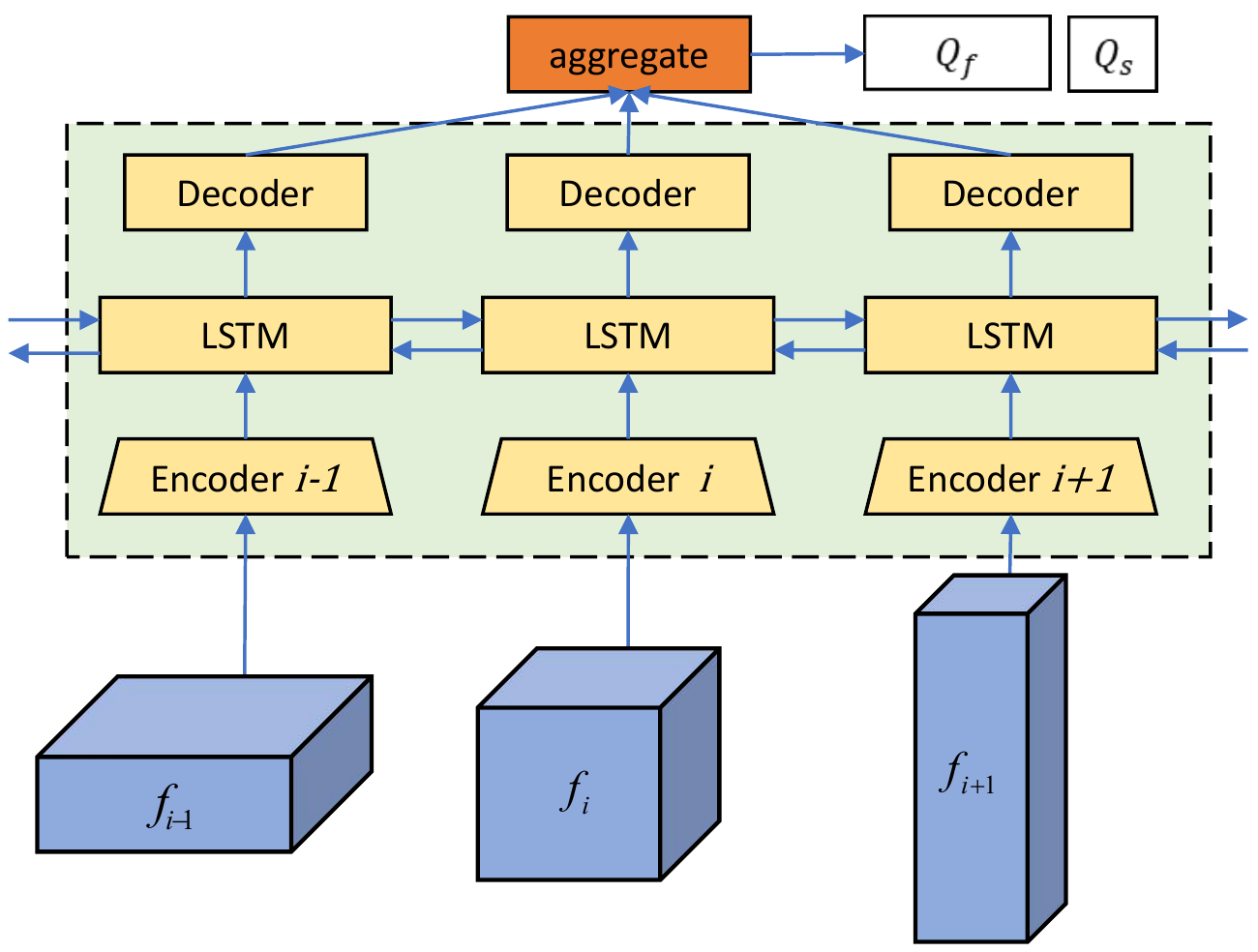}
  \vspace{5pt}
  \captionof{figure}{Q-network. It adopts the encoder-LSTM-decoder structure. The input are the current state, \ie, selected features, and the output is the predicted Q-value.}
  \label{fig:net}
\end{minipage}
\vspace{-0.3em}
\end{figure}

An overview of the proposed approach is shown in Figure~\ref{fig:arch}. Specifically, it consists of two sub-networks: a backbone CNN network which produces a collection of hierarchical features, and a decision network, \ie, the Q-network which decides how to select features produced by the backbone. The structure of the Q-network is detailed in Sec.~\ref{subsec:qnet}.

We denote the backbone CNN with $N$ convolutional layers as $C$, with each convolutional layer denoted as $C_1, C_2, \cdots, C_N$. The features produced by the layer $C_i$ ($i \in \{1,2,\cdots, N\}$) is denoted as $f_i \in \mathbb{R}^{c_i \times h_i \times w_i}$, where $c_i$ is channel size and $h_i \times w_i$ is the spatial resolution of feature $f_i$. Note that for different levels of features, the channel size and spatial size may be different. For a pair of input images, $I^s$ as the source image and $I^t$ as the target image, we denote features produced by layer $C_i$ as $f^s_i$ and $f^t_i$, respectively. The goal is to find which levels of features should be selected such that the expected matching score could be maximized. We regard the problem as a sequential decision-making problem, where at each step an agent selects a new feature map from the feature set or estimates the matching score. We apply a standard reinforcement learning setting, where each episode corresponds to a match of one image pair from a dataset. Let $\mathcal{S}$ be the state space, $\mathcal{A}$ be the set of actions and $r$ be reward functions, we instantiate $\mathcal{S}$, $\mathcal{A}$ and $\mathcal{R}$ as follows.
%
%

\head{State.} State $s = (I^s, I^t, \hat{\mathcal{F}}) \in \mathcal{S}$ at the $K$-th step consist of an image pair $(I^s, I^t)$ and the selected set of $K$ features $\hat{\mathcal{F}} = \{(f^s_i, f^t_i) \lvert i = 1, 2, \cdots, K\}$. The agent receives an observation $o=\hat{\mathcal{F}}$. In other words, the observation is the selected features of source and target images at the current step. This forms a partially observable Markov decision process, as non-selected features are invisible to the agent.

\head{Action.} The action set $\mathcal{A}$ is divided into to two subsets $\mathcal{A} = \mathcal{A}_f \cup \mathcal{A}_s$, as shown in Figure~\ref{fig:arch}. The subset $\mathcal{A}_f$ contains all actions to select new features, \ie, each action corresponds to a feature. The subset $\mathcal{A}_s$ has only one ``terminate'' action to end the episode and then compute a matching score. The number of actions in $\mathcal{A}_f$ varies based on the concrete type of the backbone CNN, \eg, $\lvert \mathcal{A}_f \rvert = 34$ for ResNet-101 as it has 34 layers. If the action $a_i \in \mathcal{A}_f$ is selected at the current step, the feature tuple $(f^s_i, f^t_i)$ will be added into $\hat{\mathcal{F}}$ and the state changes for the next step.

\head{Reward.} Each action in $\mathcal{A}_f$ indicates the corresponding feature to be selected and the agent receives a reward through this action. The reward of each action taken at the $i$-th step with the action $a_i$ is defined as:
\begin{align}
\mathcal{R}(a_i)=\begin{cases}
    -c_i, & \text{if $a_i \in \mathcal{A}_f$}\\
    \beta V_s, & \text{if $a_i \in \mathcal{A}_s$}
  \end{cases}
\enspace,
\label{eq:reward}
\end{align}
where $c_i$ is a positive cost value and $V_s$ is the final matching score. We define rewards for actions in $\mathcal{A}_f$ to be negative, and the agent gets penalty according to the total number of selected features. Driven by the reward, the agent tends to select the most discriminative features as few as possible to avoid high cost. The value $c_i$ can be either set to different values for different features $f_i$, or set to the same value for all features. Higher $c_i$ forces the agents to prefer shorter episodes, \ie, less selected features, to match images and vice versa. $\beta$ is a hyper-parameter to re-scale the matching score $V_s$ into a proper range. Usually, we expect the agent to use minimal features to achieve a matching score as high as possible, especially in cases where matching speed or hardware memory is a consideration.

\subsection{Q-network Structure}
\label{subsec:qnet}
Suppose the current state for an image pair ($I^s$, $I^t$) at step $K$ is $\hat{\mathcal{F}}=\{(f^s_1, f^t_1), \cdots, (f^s_K, f^t_K)\}$, the Q-network fuses all selected features at multiple levels to decide the next action, \ie, either to select a new feature or terminate the episode to compute a matching score. The structure of the Q-network is shown in Figure~\ref{fig:net}. It consists of an encoder-LSTM-decoder structure. Notice that features produced by different convolutional layers are of different channel dimensions and spatial resolutions. Therefore, we adopt a collection of encoders to adjust all selected feature maps to the same channel dimension and spatial resolution. Specifically, suppose the backbone CNN comprises $N$ convolutional layers, the Q-network contains $N$ separate encoders $\{E_i\}_{i=1}^N$ with the same structure but different parameter values. Each encoder processes its corresponding feature independently if the corresponding feature is selected. As features at different levels contain distinct image abstractions, each encoder can adapt its parameters to the corresponding feature level. On the contrary, if a feature at a certain level is not selected at the current step, the corresponding encoder would skip parameter updating. The structure of all encoders is: \textit{Conv - ReLU - Pooling}, where the pooling layer adaptively reduces the spatial size to $1 \times 1$ regardless of the input size.

All pooled features are fed into a bi-directional LSTM to fuse different levels of information. At each time step, the LSTM produces a latent code $L(E_i(f_i))$, regarded as embedded state information for reinforcement learning. The decoder, \ie, a fully connected layer, assimilates the latent code and produces Q-values for action decision. Note that unlike encoder networks, all decoders share the same parameters as the bidirectional LSTM integrates all contextual information at each time step. All Q-values of selected features are aggregated for the final decision. The aggregation operation used in our experiment is element-wise multiplication, and other permutation-invariant operations could also be applied as aggregation.


\subsection{Reinforcement Training}
\label{subsec:rl}
Training a deep neural network with reinforcement learning is a challenge. To stabilize and accelerate the training process, we apply the following techniques.

\head{Deep Q-learning.}~\cite{mnih2015human} proposed a combination of deep convolutional networks with a variant of Q-learning. As shown in Figure~\ref{fig:arch}, the Q-network works as a ``brain'' to make decisions. However, training a single Q-network is usually unstable in practice. We adopt a separate \textit{target network} with parameters $\theta^-$ introduced in~\cite{mnih2015human} and denote the original Q-network as \textit{evaluation network} with parameters $\theta$. The target network shares the same structure of the evaluation network, but with different parameter values. The parameters $\theta^-$ are only updated every certain steps. Instead of directly copying values of $\theta$ into $\theta^-$, we apply the method in~\cite{lillicrap2015continuous} to slowly update $\theta^-$, defined as $\theta^- := (1-\rho)\theta^- + \rho \theta$, where $\rho$ is a hyperparameter specifying the change ratio.

\head{Double Q-learning.} The Q-value at step $t$ used to guide the learning of the evaluation network is defined as:
\begin{align}
    q_t = r_t + \gamma \max_a Q^{\theta^-}(s_{t+1},a)\enspace,
    \label{eq:qeval}
\end{align}
where $\gamma$ is the discount factor for future rewards. However, as indicated in~\cite{van2016deep}, the max operator in Eq.~\ref{eq:qeval} uses the same value for action selection and evaluation, resulting in a biased estimation of the Q-value. We apply a new formula which decouples the selection and evaluation action for Q-value assignment as:
\begin{align}
    q_t = r_t + \gamma Q^{\theta^-}(s_{t+1}, \argmax_a Q^\theta(s_{t+1},a))\enspace.
    \label{eq:qdoublel}
\end{align}
In Eq.~\ref{eq:qdoublel}, the action is decided by the evaluation network but its value is estimated by the target network.

\head{Dueling Architecture.}
Following ~\cite{wang2016dueling}, our decoder consists of two parallel streams of fully-connected layers, and the latent code $L(E_i(f_i))$ produced by the LSTM is copied into two the independent streams. The first stream outputs a scalar $V(s)$ for the current state $s$, and the second stream outputs a vector $A(s, a)$ with $\lvert A \rvert$ dimensions for all actions at the state $s$. We refer readers to~\cite{wang2016dueling} for details about the decomposition of a single Q-function into two separate values. The final Q-value for an action $a$ under the state $s$ is defined as:
\begin{align}
    Q(s,a) = V(s) + A(s,a)-\frac{1}{\lvert A \rvert}\sum_{a^\prime} A(s,a^\prime ) \enspace.
\end{align}
By estimating the value of state $V(s)$ explicitly, the training process speeds up and stabilizes.

\head{Retrace.}
As we store trajectories into a memory buffer and randomly sample trajectories during each training iteration, there exists a discrepancy between the sampled trajectories and the current policy. To resolve the discrepancy, we modify the Q-value at step $t$ based on the new estimation proposed in~\cite{janisch2019classification}:
\begin{align}
\begin{split}
    q_t=r_t + \gamma \mathbb{E}_{a\sim \pi (s_t)} [Q(s_{t+1}, a)] +
    \gamma \bar{\rho}_{t+1}[q_{t+1}-Q(s_{t+1},a_{t+1})]\enspace,
\end{split}
\end{align}
where $\bar{\rho}_t = \min(\frac{\pi(a_t|s_t)}{\mu(a_t|s_t)}, 1)$ is a truncated importance sampling between behavior policy $\mu$ that is used to generate trajectories stored in the memory buffer and the target policy $\pi$ that the Q-network aims to learn. Importance sampling is a simple way to correct the discrepancy between $\mu$ and $\pi$ when off-policy learning, \ie, using memory buffer, is applied.

\section{Experiments}
In this section, we present a comprehensive implementation, evaluation, and analysis of our proposed approach on three public real image datasets.

\subsection{Datasets and Metric}
We compare our model to other image matching methods with both hand-crafted and CNN-based features on three datasets: PF-PASCAL~\cite{ham2017proposal}, PF-WILLOW~\cite{ham2017proposal} and Caltech-101~\cite{fei2006one}.

\head{Datasets.} The PF-PASCAL dataset consists of 1,351 semantically related image pairs from 20 object categories. 
All image pairs are split into training, validation, and test sets in~\cite{han2017scnet}. Manually annotated correspondences for each pair are only provided in validation and test sets. As we need matching scores as the reward of the terminal state, we further split the original validation set into two parts with a ratio of $8:2$ as the new training and validation sets. Therefore, we use much less images for training. The PF-WILLOW dataset comprises 100 images which are grouped into 900 image pairs. All pairs are divided into four semantically related subsets. For each image, 10 keypoint annotations are provided. 
The Caltech-101 dataset provides images of 101 object categories with ground-truth object masks, but 
it does not provide ground-truth keypoint annotations. Following~\cite{rocco2018end}, we choose 15 image pairs for each object category and use the corresponding 1,515 image pairs for evaluation.

\head{Metric.} We adopt the percentage of correct keypoints (PCK) metric to evaluate our model on PF-PASCAL and PF-WILLOW. PCK measures the percentage of keypoints whose transformation errors are below a given threshold. The transformation error is measured as the Euclidean distance between the location of a warped keypoint and its corresponding ground-truth keypoint. The threshold is defined as $\alpha\max(h,w)$ where $h$ and $w$ are height and width of the object bounding box. 
For both datasets, we set the threshold $\alpha=0.1$. Following~\cite{kim2013deformable}, we adopt label transfer accuracy (LT-ACC) and intersection-over-union (IoU) to evaluate the performance on Caltech-101 dataset. Both metrics measure the number of correctly labeled pixels between ground-truth and warped masks generated by estimated correspondences.

\subsection{Implementation and Training}
We adopt ResNet-101 pre-trained on ImageNet~\cite{deng2009imagenet} classification task as the backbone network. In our Q-network, the number of encoders is 34, as ResNet-101 contains 34 convolutional layers. The convolutional layers in all encoders 
output a 512-dimensional feature map. The hidden size of the LSTM and the fully-connected layer size in the decoder are also set to be 512. We use features produced by the fully-connected layer of the backbone as initial states to start selection episodes, and these initial features are not applied in computing matching scores. Once the action in $\mathcal{A}_s$ is chosen, we adopt the probabilistic Hough matching proposed in~\cite{min2019hyperpixel} to match images using selected features. During training, we fix the parameters of the backbone network and only train the Q-network. The costs for all actions in $\mathcal{A}_f$ are fixed to 0.4 and $\beta$ is set to 20. We empirically set the initial learning rate to 0.001. We adopt the Adam optimizer for optimization with $\beta_1 = 0.9$ and $\beta_2 = 0.999$. We train our model with a batch size of 16 for 3000 iterations and apply early-stopping. We keep track of the average reward on the training set and decrease the learning rate immediately once the reward fails to increase for ten times. The average reward on the validation set is used for early stopping if the reward fails to increase consecutively for twenty times.

\begin{table}[!htb]
\small
\centering
\setlength{\tabcolsep}{1.5pt}
\begin{minipage}{.54\linewidth}
    \centering
\begin{tabular}{l|l|c|c}
\multirow{2}{*}{}             & \multicolumn{1}{c|}{\multirow{2}{*}{Model}}       & \multicolumn{2}{c}{PCK ($\alpha=0.1$)} \\
                              & \multicolumn{1}{c|}{}                             & \multicolumn{1}{c|}{PF-WILLOW}      & PF-PASCAL     \\ \hline\hline
\multirow{5}{*}{\rotatebox[origin=c]{90}{Hand-crafted}} & DeepFlow~\cite{revaud2016deepmatching}           & \multicolumn{1}{c|}{0.20}           & 0.21          \\
                              & GMK~\cite{duchenne2011graph}                 & \multicolumn{1}{c|}{0.27}           & 0.27          \\
                              & DSP~\cite{kim2013deformable}                   & \multicolumn{1}{c|}{0.29}           & 0.30          \\
                              & SIFTFlow~\cite{liu2010sift}            & \multicolumn{1}{c|}{0.38}           & 0.33          \\
                              & ProposalFlow~\cite{ham2017proposal}   & \multicolumn{1}{c|}{0.56}           & 0.45          \\ \hline
\multirow{6}{*}{\rotatebox[origin=c]{90}{CNN-based}}    &  FCSS + PF-LOM~\cite{kim2017fcss}                   & \multicolumn{1}{c|}{0.58}           & 0.46          \\
                              & GeoCNN (SS)~\cite{rocco2017convolutional} & \multicolumn{1}{c|}{0.68}               & 0.68              \\
                              & A2Net~\cite{hongsuck2018attentive}     & \multicolumn{1}{c|}{0.69}               & 0.67              \\
                              & GeoCNN (WS)~\cite{rocco2018end}   & \multicolumn{1}{c|}{0.71}           & 0.72          \\
                              & SFNet~\cite{lee2019sfnet}   & \multicolumn{1}{c|}{\underline{0.74}}          & 0.79    \\
                              & HPFlow~\cite{min2019hyperpixel}   & \multicolumn{1}{c|}{\underline{0.74}}      & \underline{0.85}    \\
                              & Ours                                              & \multicolumn{1}{c|}{\textbf{0.75}}               & \textbf{0.86}              \\ \hline
\end{tabular}
\vspace{8pt} 
\caption{The average PCK results on PF-WILLOW and the test split of PF-PASCAL dataset with $\alpha = 0.1$. Numbers of the top-1 performance are in bold and the top-2 performance are underlined.}
\label{tab:pf}
\end{minipage}\enspace\enspace%
\begin{minipage}{.38\linewidth}
    \centering
\begin{tabular}{l|l|c|c}
\multirow{2}{*}{}                  & \multicolumn{1}{c|}{Methods} & LT-ACC & IoU  \\ \hline\hline
\multirow{6}{*}{\rotatebox[origin=c]{90}{Hand-crafted}} & DeepFlow~\cite{revaud2016deepmatching}                    & 0.74   & 0.40 \\
                      & SIFTFlow~\cite{liu2010sift}                     & 0.75   & 0.48 \\
                      & GMK~\cite{duchenne2011graph}                          & 0.77   & 0.40 \\
                      & DSP~\cite{kim2013deformable}                          & 0.77   & 0.47 \\
                      & ProposalFlow~\cite{ham2017proposal}                       & 0.78   & 0.50 \\
                      & OADSC~\cite{yang2017object}                        & 0.81   & 0.55 \\ \hline
\multirow{7}{*}{\rotatebox[origin=c]{90}{CNN-based}}  & SCNet-AG\cite{han2017scnet}                    & 0.79   & 0.51 \\
                      & A2Net~\cite{hongsuck2018attentive}                        & 0.80   & 0.57 \\
                      & FCSS + PF-LOM~\cite{kim2017fcss}                         & 0.83   & 0.52 \\
                      & GeoCNN (SS)~\cite{rocco2017convolutional}                       & 0.83   & 0.61 \\
                      & GeoCNN (WS)~\cite{rocco2018end}                       & 0.85   & \underline{0.63} \\
                      & SFNet~\cite{lee2019sfnet}                        & \textbf{0.88}   & \textbf{0.67} \\
                      & HPFlow~\cite{min2019hyperpixel}   & \multicolumn{1}{c|}{\underline{0.87}}      & \underline{0.63}    \\
                      & Ours                         & \underline{0.87}   & \underline{0.63} \\ \hline
\end{tabular}
\vspace{8pt} 
\caption{The average quantitative results on Caltech-101 dataset. Numbers of the top-1 performance are in bold and the top-2 performance are underlined.}
\label{tab:cal}
\end{minipage}
\vspace{-3pt}
\end{table}

\begin{figure*}[ht]
\captionsetup[subfigure]{labelformat=empty}
\centering
\begin{tabular}{c @{\hspace{0.1\tabcolsep}}c@{\hspace{0.1\tabcolsep}}c@{\hspace{0.1\tabcolsep}}c@{\hspace{0.1\tabcolsep}}c@{\hspace{0.1\tabcolsep}}c}
\subfloat{\includegraphics[width = 0.16\linewidth]{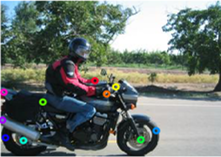}} &
\subfloat{\includegraphics[width = 0.16\linewidth]{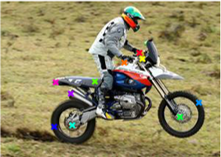}} &
\subfloat{\includegraphics[width = 0.16\linewidth]{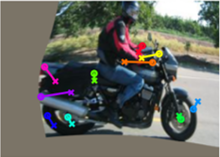}} & 
\subfloat{\includegraphics[width = 0.16\linewidth]{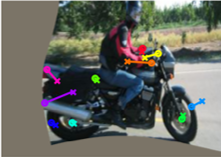}} &
\subfloat{\includegraphics[width = 0.16\linewidth]{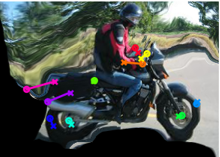}} &
\subfloat{\includegraphics[width = 0.16\linewidth]{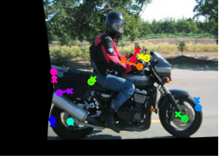}} \\ [-2.75ex]
\subfloat{\includegraphics[width = 0.16\linewidth]{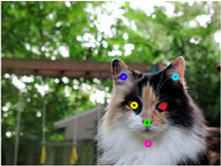}} &
\subfloat{\includegraphics[width = 0.16\linewidth]{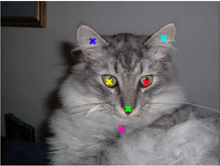}} &
\subfloat{\includegraphics[width = 0.16\linewidth]{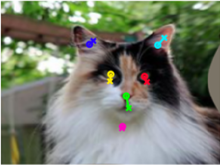}} &
\subfloat{\includegraphics[width = 0.16\linewidth]{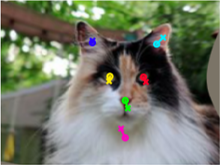}} &
\subfloat{\includegraphics[width = 0.16\linewidth]{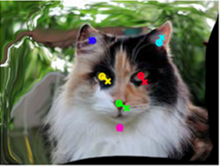}} &
\subfloat{\includegraphics[width = 0.16\linewidth]{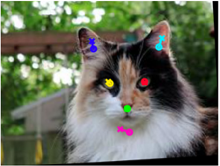}} \\ [-2.75ex]
\subfloat{\includegraphics[width = 0.16\linewidth]{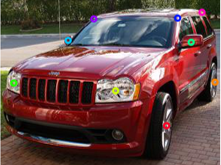}} &
\subfloat{\includegraphics[width = 0.16\linewidth]{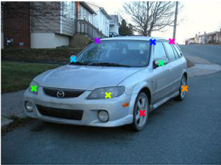}} &
\subfloat{\includegraphics[width = 0.16\linewidth]{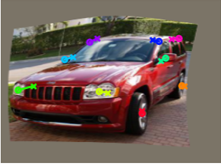}} &
\subfloat{\includegraphics[width = 0.16\linewidth]{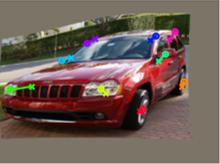}} &
\subfloat{\includegraphics[width = 0.16\linewidth]{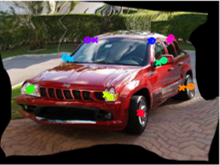}} &
\subfloat{\includegraphics[width = 0.16\linewidth]{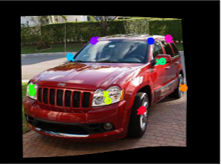}} \\ [-2.75ex]
\subfloat{\includegraphics[width = 0.16\linewidth]{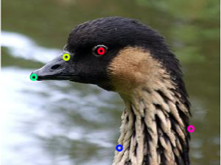}} &
\subfloat{\includegraphics[width = 0.16\linewidth]{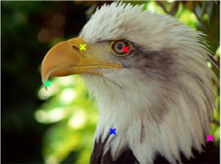}} &
\subfloat{\includegraphics[width = 0.16\linewidth]{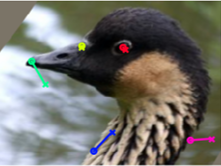}} &
\subfloat{\includegraphics[width = 0.16\linewidth]{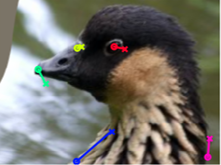}} &
\subfloat{\includegraphics[width = 0.16\linewidth]{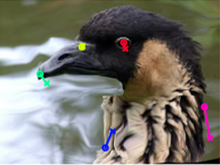}} &
\subfloat{\includegraphics[width = 0.16\linewidth]{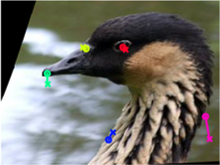}} \\ [-2.75ex]
\subfloat[Source image]{\includegraphics[width = 0.16\linewidth]{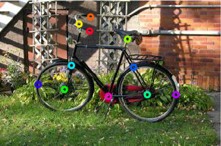}} &
\subfloat[Target image]{\includegraphics[width = 0.16\linewidth]{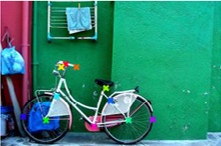}} &
\subfloat[GeoCNN (WS)]{\includegraphics[width = 0.16\linewidth]{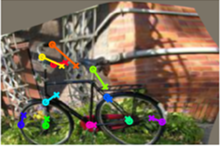}} &
\subfloat[A2Net]{\includegraphics[width = 0.16\linewidth]{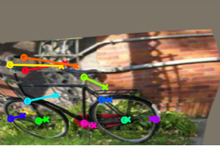}} &
\subfloat[SFNet]{\includegraphics[width = 0.16\linewidth]{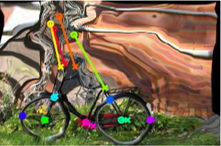}} &
\subfloat[Ours]{\includegraphics[width = 0.16\linewidth]{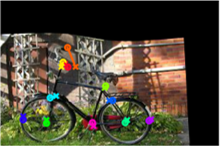}}
\end{tabular}
\vspace{5pt}
\caption{Examples of qualitative results from PF-PASCAL dataset. Keypoints of the source and target images are shown in circles and crosses, respectively. Compared to GeoCNN (WS)~\protect\cite{rocco2018end}, A2Net~\protect\cite{hongsuck2018attentive} and SFNet~\protect\cite{lee2019sfnet}, our method is more robust to intra-class variations. 
}
\label{fig:qualitative}
\end{figure*}

\subsection{Results}
\head{PF-WILLOW \& PF-PASCAL.} In Table~\ref{tab:pf}, we report the average PCK scores of our method and recent methods that are directly comparable. 
Note that the performance of all compared methods are taken from~\cite{ham2017proposal,hongsuck2018attentive,lee2019sfnet,min2019hyperpixel}. As shown in Table~\ref{tab:pf}, our proposed method achieves state-of-the-art performance~\cite{min2019hyperpixel}. Our model is trained on PF-PASCAL dataset and the selected features are produced by (2, 23, 25, 28) layers, which is around half of all selected layers proposed by~\cite{min2019hyperpixel}\footnote{The layers selected for PF-PASCAL dataset using ResNet-101 in~\cite{min2019hyperpixel} are (2, 17, 21, 22, 25, 26, 28).}. For PF-WILLOW dataset, in order to verify the generalization ability of our proposed method, we directly apply the same selected layers for testing without fine-tuning. The results in Table~\ref{tab:pf} indicates: 1) fusing multiple levels of features is beneficial for image matching; 2) based on our instantiation of three core components of reinforcement learning, \ie, state, action and reward, our method has the capability to learn an optimal selection policy for image matching. As we assign each selected feature a positive cost, it drives our method to select features as few as possible, trading off between the final reward and the cost to receive that reward. Another possible reason leading to fewer selected features and higher scores is that there is no restriction for selection order during the policy learning process. On the contrary, beam search in~\cite{min2019hyperpixel} selects features from low levels to high levels sequentially without skipping, resulting in a smaller search space.

\head{Caltech-101.} The quantitative results for the Caltech-101 dataset are listed in Table~\ref{tab:cal}. All results except for ours are taken from~\cite{ham2017proposal,hongsuck2018attentive,lee2019sfnet,min2019hyperpixel}. Similar to PF-WILLOW dataset, we directly test our method on Caltech-101 using selected layers based on PF-PASCAL without fine-tuning. Our method achieves the same performance as~\cite{min2019hyperpixel}. Notice that the performance of SFNet~\cite{lee2019sfnet} is better than ours and~\cite{min2019hyperpixel}. One possible explanation is that we select features from the backbone network pre-trained for image classification, while SFNet introduces two trainable convolutional layers to refine image features for image matching. Although we combine both high and low levels of features, the combined features are still not as robust as features dedicatedly trained for image matching. Therefore, incorporating feature learning into our proposed method is left for future study.

\head{Qualitative comparison.} Figure~\ref{fig:qualitative} visualizes the matching results between keypoints in source and target images on the test split of the PF-PASCAL dataset. We can see that our method is able to select features that are robust to (near) rigid deformations (\eg, cars in the third row), local non-rigid deformations (\eg, the cat's head in the second row and the upper part of birds in the fourth row), scale changes between objects (\eg, wheels of bicycles in the last row). In particular, the first example clearly demonstrates that our method establishes more discriminative correspondences, avoiding matches for non-target objects. For example, it does not match keypoints on the motorbike to keypoints on the rider between the source and target images, while all other methods mismatch some keypoints.

\section{Conclusion}
We propose a reinforcement learning method to dynamically select an optimal set of features at multiple levels to match images. Our method regards the collection of selected features as states and simulates an environment to generate rewards to decide actions to select the next feature. The experiments on three public benchmarks demonstrate that our method is able of selecting a small set of features for image matching. We believe applying reinforcement learning to select task-relevant visual features can be applied to various vision tasks.

\section*{Appendix}
\label{sec:appendix}
\head{A.1} In this subsection, we briefly summarize the process of establishing keypoint correspondences using probabilistic Hough matching~\cite{min2019hyperpixel}. Given a collection of selected features, firstly, all selected features are spatially resized to the same size as the largest feature and then concatenated along the channel dimension. Given any two positions in source and target image features separately, we compute their visual similarity (cosine distance) and the spatial offset between these two positions. A Hough space is built based on offsets and discretized into different bins. Each pair of positions in source and target features cast a vote, \ie, the value of visual similarity, to their corresponding offset bin. The final visual similarity between any two positions is re-weighted by summing up all votes falling into their bin.
\begin{wraptable}[15]{r}{0.5\linewidth}
\footnotesize
\centering
\begin{tabular}{c|l|c}
\# Layers          & \multicolumn{1}{c|}{Layer Indices} & PCK  \\ \hline\hline
\multirow{2}{*}{2} & (4, 6)                           & 0.05 \\
                  & (6, 18)                          & 0.11 \\ \hline
\multirow{2}{*}{3} & (17, 22, 31)                     & 0.70 \\
                  & (14, 27, 30)                     & 0.71 \\ \hline
\multirow{2}{*}{4} & (7, 19, 30, 31)                  & 0.10 \\
                  & (9, 15, 16, 26)                  & 0.59 \\ \hline
\multirow{2}{*}{5} & (6, 14, 17, 18, 22)              & 0.38 \\
                  & (9, 16, 17, 21, 24)              & 0.66 \\ \hline
\multirow{2}{*}{6} & (5, 7, 14, 27, 32, 33)           & 0.17 \\
                  & (9, 15, 22, 23, 25, 33)          & 0.73 \\ \hline
\multirow{2}{*}{7} & (6, 10, 11, 20, 28, 29, 33)      & 0.25 \\
                  & (16, 18, 22, 26, 27, 30, 33)     & 0.70 \\ \hline
\end{tabular}
\vspace{1em} 
\caption{The average PCK results of randomly selected layers on PF-PASCAL dataset.}
\label{tab:random}
\end{wraptable}
The keypoint correspondences are estimated as follows: for a keypoint in a source image, there must exist several receptive fields (RFs, corresponding to positions in image features) covering it. Each position in the source feature has a most visually similar position in the target feature corresponding to an RF in the target image. The displacements of source RFs' centers to the source keypoint are used as weights to weighted sum up the corresponding target RFs' centers, which is regarded as the source keypoint's correspondence.

\head{A.2} In this subsection, we report the experiment on the PF-PASCAL dataset by randomly selecting a collection of layers. We fix the number of layers to be selected, \ie, the value of $K$ is restricted within the range from 2 to 7, and then randomly selected $K$ layers. For each value of $K$, we test twice and the result is shown in Table~\ref{tab:random}. Notice that even if the same number of $K$ layers are used for matching, the randomly selected layers yield inferior performance compared with layers selected by the proposed method, as shown in Table~\ref{tab:pf}.

\section*{Acknowledgement}
We would like to thank the reviewers for their comments and efforts towards improving our manuscript. The authors appreciate the generous support provided by Inception Institute of Artificial Intelligence in the form of NYUAD Global Ph.D. Student Fellowship.

\bibliography{egbib}
\end{document}